\documentclass[conference]{IEEEtran}
\IEEEoverridecommandlockouts
\usepackage{cite}
\usepackage{amsmath,amssymb,amsfonts}
\usepackage{algorithmic}
\usepackage{graphicx}
\usepackage{textcomp}
\usepackage{xcolor}
\usepackage{subcaption}
\usepackage[hidelinks]{hyperref}
\usepackage{enumitem}
\usepackage{amsmath}
\usepackage{multirow} 
\usepackage{array}
\usepackage{url}
\usepackage{float}

\newcommand{\PreserveBackslash}[1]{\let\temp=\\#1\let\\=\temp}
\newcolumntype{C}[1]{>{\PreserveBackslash\centering}p{#1}}

\def\BibTeX{{\rm B\kern-.05em{\sc i\kern-.025em b}\kern-.08em
    T\kern-.1667em\lower.7ex\hbox{E}\kern-.125emX}}
\begin{document}

\makeatletter
\newcommand{\linebreakand}{%
  \end{@IEEEauthorhalign}
  \hfill\mbox{}\par
  \mbox{}\hfill\begin{@IEEEauthorhalign}
}
\makeatother

\title{Socially Aware Motion Planning for Service Robots Using LiDAR and RGB-D Camera}

\author{
\IEEEauthorblockN{1\textsuperscript{st} Duc Phu Nguyen}
\IEEEauthorblockA{\textit{Fulbright University Vietnam}\\
Ho Chi Minh City, Vietnam \\
phu.nguyen.200131@student.fulbright.edu.vn}
\and
\IEEEauthorblockN{2\textsuperscript{nd} Thanh Long Nguyen}
\IEEEauthorblockA{\textit{Fulbright University Vietnam}\\
Ho Chi Minh City, Vietnam \\
long.nguyen.210085@student.fulbright.edu.vn}
\and
\IEEEauthorblockN{3\textsuperscript{rd} Minh Dang Tu}
\IEEEauthorblockA{\textit{Vietnam National University}\\
Hanoi, Vietnam\\
dangyuuki@gmail.com}
\linebreakand
\IEEEauthorblockN{4\textsuperscript{th} Cong Hoang Quach}
\IEEEauthorblockA{\textit{University of Technology Sydney}\\
Sydney, Australia \\
conghoang.quach@uts.edu.au}
\and
\IEEEauthorblockN{5\textsuperscript{th} Xuan Tung Truong}
\IEEEauthorblockA{\textit{Le Quy Don University}\\
Hanoi, Vietnam \\
tungtx@lqdtu.edu.vn}
\and
\IEEEauthorblockN{6\textsuperscript{th} Manh Duong Phung}
\IEEEauthorblockA{\textit{Fulbright University Vietnam}\\
Ho Chi Minh City, Vietnam \\
duong.phung@fulbright.edu.vn}
}

\graphicspath{{images/}}
\maketitle
\begin{abstract}
Service robots that work alongside humans in a shared environment need a navigation system that takes into account not only physical safety but also social norms for mutual cooperation. In this paper, we introduce a motion planning system that includes human states such as positions and velocities and their personal space for social-aware navigation. The system first extracts human positions from the LiDAR and the RGB-D camera. It then uses the Kalman filter to fuse that information for human state estimation. An asymmetric Gaussian function is then employed to model human personal space based on their states. This model is used as the input to the dynamic window approach algorithm to generate trajectories for the robot. Experiments show that the robot is able to navigate alongside humans in a dynamic environment while respecting their physical and psychological comfort.
\end{abstract}

\section{Introduction}
An effective navigation system for service robots goes beyond just obstacle avoidance as the robots work alongside humans in a dynamic environment with social interaction. The system should anticipate pedestrian flow, allowing the robot to navigate seamlessly without being intrusive. Imagine a delivery robot in a hospital; it should take a route that minimizes disruption to patients and staff while still completing its task efficiently. In this context, the disruption can be divided into two categories: (i) physical disruption and (ii) psychological disruption. The first category requires the robot to maintain a minimum physical distance from humans. The second category, on the other hand, involves a personal space around a pedestrian, to which intrusion might cause discomfort to the person \cite{8011466}. 

To address the physical safety problem, traditional approaches can be applied in which humans are considered as dynamic obstacles so that obstacle avoidance techniques can be used. Depending on the sensors used, the techniques can vary from the artificial potential field (APF) \cite{rostami2019obstacle} and dynamic window approach (DWA) \cite{fox1997dynamic} to randomized kinodynamic planning \cite{lavalle2001randomized} and reciprocal
velocity obstacles (RVOs) \cite{van2008reciprocal}. These techniques were designed to avoid physical collisions between the robot and the human. However, they do not take into account human characteristics and social context to address psychological safety.

To maintain both physical and psychological comfort, it is necessary to include human state and social behavior norms in the algorithms. Approaches to this problem can be categorized into the model-based approach \cite{8011466}, and the learning-based approach \cite{chen2017socially}. In the model-based approach, the personal space of a person is modeled by a cost function. In \cite{kirby2010social}, an asymmetric Gaussian function is used. In \cite{papadakisadaptive}, non-stationary, skew-normal probability density functions are developed for social mapping considering social cues, certainty, and robot perception capacity. On a larger scale, a group of people interacting with each other is mapped in \cite{papadakis2013social, truong2016dynamic}. These models are incorporated into the path-planning algorithm to avoid potential collision and causing discomfort to humans. The learning-based approach, on the other hand, does not use a specific model of humans but learns it via a large number of trials. Deep reinforcement learning is often used to generate navigation policies that respect humans' personal space \cite{chen2017socially, 2043184}. This approach, however, requires a large amount of data and high computational cost, which may not be suitable for embedded computers used in the robot.

In social-aware motion planning, the robot needs to have a sensory system sufficient to detect and track humans. Commonly used sensors include LiDAR \cite{nakamori2018multiple} and camera \cite{liu2016people}. While LiDAR excels at measuring distance and direction, it is error-prone to mistake humans for other objects, especially in clustered environments \cite{sarmento2024fusion}. The camera, on the other hand, can provide richer data for human detection. It however has a limited field of view which causes problems for the robot to observe its surrounding environment. Using both LiDAR and camera thus is desirable for human detection and navigation \cite{chebotareva2020person}.

In this paper, we propose a motion planning system that takes into account social context to generate trajectories for the robot. The trajectories maintain both physical and psychological safety when operating in dynamic environments with humans involved. The system includes person-tracking algorithms for LiDAR and the RGB-D camera and a motion planning algorithm using the data fused from these two observation sources for reliable navigation planning. The contributions of this study are threefold: (i) the development of a robust human state estimation algorithm using color and depth information of the RGB-D camera; (ii) the proposal of a method to fuse data from LiDAR and the RGB-D camera for state estimation; (iii) implementation of a motion planner for the robot that considers humans' personal space for social-aware navigation.

\section{Human model and robot sensor system}\label{sec:bg}
This section describes the human model and the sensors of the robot used in our socially aware motion planning system.   

\begin{figure}[h]
\centering
\begin{subfigure}[t]{0.45\linewidth}
    \centering
    \includegraphics[width=1\textwidth]{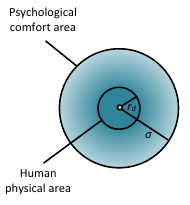}
    \caption{}
    \label{fig:pstatic}
\end{subfigure}
\hspace{0.05cm}
\begin{subfigure}[t]{0.51\linewidth} 
    \centering
    \includegraphics[width=1\textwidth]{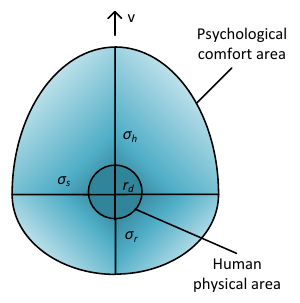}
    \caption{}
    \label{fig:pmoving}
\end{subfigure}
\caption{Personal space: (a) static person; (b) moving person}
\label{fighumanmodel}
\end{figure} 

\subsection{Human model}
When a robot operates in a shared environment, it needs to have the capability to not only avoid physical collisions with humans but also respect their personal space for mutual cooperation. We address this by modeling a human with two zones, the physical and psychological comfort areas, as shown in Fig. \ref{fighumanmodel}. The physical area represents the human body and is modeled as a circle with radius $r_d$. The psychological comfort area represents the personal space that the robot needs to respect to avoid interruption and discomfort to humans' activities. It is modeled by a Gaussian function with the variance dependent on human states. The states are divided into two classes, stationary and moving. A static person is then modeled with a fixed variance $\sigma$ resulting in a symmetric Gaussian function with a round shape, as shown in Fig. \ref{fig:pstatic}. 

When a person moves with a velocity $v_p$ and heading direction $\theta_p$, the personal space is modeled as an asymmetric Gaussian function of an egg shape with its front side aligned with the heading direction, as shown in Fig. \ref{fig:pmoving}. The variances $\sigma_h$, $\sigma_s$, and $\sigma_r$ of the Gaussian function representing the heading, side, and rear directions of the person are respectively chosen as follows \cite{kirby2010social}:     

\begin{gather}
    \sigma_h = max(2v_p, \frac{1}{2}) \\
    \sigma_s = \frac{2}{3} \sigma_h \\
    \sigma_r = \frac{1}{2} \sigma_h
\end{gather} 

In our work, we implement this model using a social costmap layer based on \cite{lu2014layered}. The size of the personal space in the costmap can be modified by adjusting the cost function's amplitude and variance values, to which a scaling factor to the velocity can be added. This allows tuning the planned path to move closer or further to the person to acquire desirable path-planning behavior. 

\begin{figure}[h]
\centering
\includegraphics[width=0.4\textwidth]{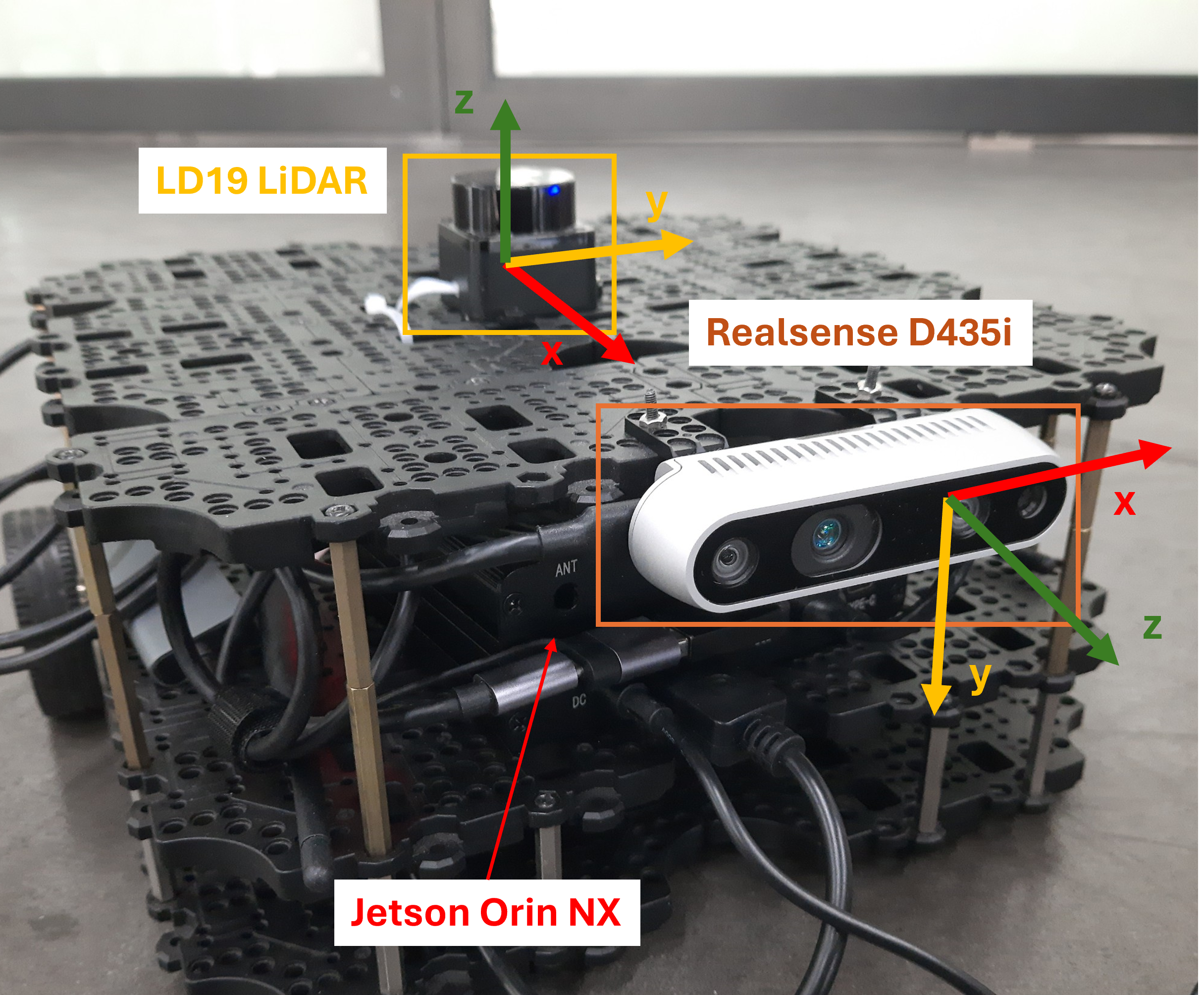}    
\caption{The robot platform with LiDAR and RGB-D camera}
\label{fig:robot}
\end{figure} 

\subsection{Robot sensor system}
To detect and localize humans in a complex environment for social-context motion planning, the robot needs a robust and reliable sensory system. We propose in this work the use of a LiDAR sensor and an RGB-D camera. The LiDAR, with $360^o$ scanning range, accurate distance measurements, and a fast sampling rate, provides real-time data of the surrounding environment for mapping, localization and navigation. The camera, on the other hand, can provide rich visual details like color and texture for human detection. In addition, its depth channel enables data association between the camera and the LiDAR. Combining these sensors hence can leverage their strengths for more robust human detection and navigation. 

In this work, the LiDAR used is a DTOF LD19 mounted at the center of the robot, as shown in Fig. \ref{fig:robot}. The camera is an Intel Realsense D435i mounted at the front of the robot and has a field of view (FoV) of $87^o$. Data from these sensors is processed by a single-board computer, Jetson Orin NX. 
\subsection{LiDAR and Camera calibration}
Since the LiDAR coordinate is aligned with the robot coordinate, we need to convert the camera coordinate to the LiDAR coordinate for data fusion. Denote a point in the camera's frame as $(x_C, y_C, z_C)$ and its correspondence in the LiDAR's frame as $(x_L, y_L, z_L)$. Note that $z_L = 0$ as the LiDAR does not include height information. The homogeneous transformation from the camera coordinate to the LiDAR coordinate is then given by:

\begin{equation}
    \begin{bmatrix}
        x_{L} \\
        y_{L} \\
        0 \\ 
        1
    \end{bmatrix} =
    \begin{bmatrix}
        0 && 0 && 1 && \Delta x \\
        1 && 0 && 0 && \Delta y \\
        0 && 0 && 0 && 0 \\
        0 && 0 && 0 && 1
    \end{bmatrix} \begin{bmatrix}
        x_C \\
        y_C \\
        z_C \\
        1
    \end{bmatrix},
    \label{eq:transform}
\end{equation}
where $\Delta x$ and $\Delta y$ are the translations of the camera with respect to the LiDAR.

To determine $\Delta x$ and $\Delta y$, we placed a box with an ArUco marker in front and at the center of the robot. An algorithm is then used to detect the marker and calculate its coordinate $(x_a, y_a, z_a)$ in the camera's frame. The distance $d$ from the robot center to the box can be extracted from the LiDAR data. The translations then can be calculated as follows:
\begin{gather}
    \Delta x = d - z_a \\
    \Delta y = x_a
\end{gather}

\section{Human state estimation} 
This section describes the detection of human states from LiDAR and the RGB-D camera and our approach to fuse those states to give a more robust and reliable estimation. 

\subsection{Human detection using LiDAR}
The detection of humans from LiDAR is conducted using the Joint Leg Tracker framework \cite{leigh2015person}. This framework first clusters data points based on their relative distance. The clusters are then classified as human or non-human by using a random forest classifier that learns geometric features from the data. The framework returns the human positions together with their standard deviations that can be used for the fusion algorithm.

\subsection{Human detection using the RGB-D camera}
The proposed approach employs a combination of a pose estimator and depth frames to determine the human's position in the robot frame. 
\subsubsection{Pose estimator with MoveNet}
For human pose estimation, the selected framework is the Multipose Lighting MoveNet developed by Google \cite{movenet}. This choice is motivated by its lightweight architecture, which aligns well with the real-time demands of embedded systems such as the Jetson Orin. The output of MoveNet is a set of keypoints of the human body, such as the hip, knee, and ankle, which can be used to determine the human's position. Note that since the camera is aligned with the robot chassis, its field of view is often limited to the lower part of the body, as shown in Fig. \ref{fig:depth_camera}.

\subsubsection{Human's position in the LiDAR frame}
To determine the human's position in the LiDAR frame, we first calculate that position in the camera frame based on the keypoints obtained from the pose estimator. The keypoints are prioritized based on their inherent stability in the following order: (i) hip, (ii) knee, and (iii) ankle. This prioritization is based on the anatomical structure of the human body where hips exhibit less movement variability compared to knees and ankles, leading to more accurate position estimation. Denote $L_i$ and $R_i$ respectively as the left and right keypoints of join $i$. The position of that join is given by
\begin{equation}
J_i = \frac{L_i + R_i}{2}.
\end{equation}

Given a set of join positions $J_i$, the join with the highest priority, $J^*_i$, is chosen to represent the human's position. The depth channel is then used to identify the depth value at position $J^*_i$. Denote $(x_J, y_J, z_J)$ as the coordinates of $J^*_i$ in the camera frame. Its coordinate in the LiDAR frame then can be computed by multiplying it with the transformation matrix defined in (\ref{eq:transform}). 

\subsection{Fusion algorithm for human state estimation}
Given the human positions obtained from the LiDAR and the camera, it is necessary to fuse them to get a final stable position. We use the Kalman Filter for this task. The human states we want to track include the position $(x, y)$ and velocity $(\dot{x}, \dot{y})$. Denote 
$\mathbf{x}_k = [x, y, \dot{x}, \dot{y}]^T$ as the state vector at time step $k$. We use the constant velocity model to describe the human movement as follows: 
\begin{gather}
    \mathbf{x}_{k} = A \mathbf{x}_{k-1} + B u_{k-1} + w_{k-1},
\end{gather} 
where $\mathbf{x}_{k-1}$ is the previous (posterior) state, $A$ is the state transition matrix, $B$ is the control input matrix, and $w_{k-1}$ is the process noise having a normal distribution with zero mean and variance $Q$, $p(w) \sim N(0, Q)$.

The system has two observations of the human state, acquired from the LiDAR and the camera. We denote them as $z_{k1} = Hx_k + v_{k1}$ and $z_{k1} = Hx_k +v_{k2}$, where $H$ is the observation matrix, and $v_{k1}, v_{k2}$ are the measurement noises of two measurements with the normal probability $p(v_{k1}) \sim N(0, R_1)$ and $p(v_{k2}) \sim N(0, R_2)$, respectively. We assume $R_1$ and $R_2$ are the constant measurement noise covariance and independent of each other.    
At each time step $k$, the Kalman filter receives the previous state $x_{k-1}$ and the estimated error covariance $P_{k-1}$. Then, the filter predicts the priori state $\mathbf{x}_{k}^{(P)}$ as 
\begin{gather}
    \mathbf{x}_{k}^{(P)} = A \mathbf{x}_{k-1} + B u_{k-1} 
\end{gather} 
and the priori error covariance as
\begin{gather}
    P_{k}^{(P)} = A P_{k-1} A^{T} + Q.
\end{gather} 

Since the system has two observation sources, we use an asynchronous approach in which the state is updated with whichever measurement comes first. The process for updating with a measurement $z_k$ having the corresponding measurement noise covariance $R$ is as follows:

The filter computes the Kalman gain
\begin{gather}
    K_{k} = P_{k}^{(P)}H^{T}(HP_{k}^{(P)}H^{T} + R)^{-1} 
\end{gather}
and then updates the posteriori state
\begin{gather}
    x_{k} = {x}_{k}^{(P)} + K_{k}(z_{k}-H{x}_{k}^{(P)}).
\end{gather} 
Lastly, the filter updates a posteriori error covariance 
\begin{gather*}
    P_{k} = (I - K_{k}H)P_{k}^{(P)},
\end{gather*} where $I$ is the identity matrix.

\section{Socially aware motion planning}
In this work, we employ a costmap-based navigation framework for the robot. The fused human states over time are mapped into a local costmap layer to represent their social behaviors. This local costmap layer is then updated to the master costmap for navigation. The dynamic window approach (DWA) algorithm \cite{fox1997dynamic} is then used for motion planning. The algorithm finds a pair of linear and angular velocities describing the optimal trajectory to which the robot can travel in the next update. 

The robot in this work is a differential drive robot with the kinematic model in the discrete-time domain given as follows:  
\begin{gather}
    x_{k+1} = x_{k} + v_k\Delta t cos(\theta_k) \\
    y_{k+1} = y_{k} + v_k\Delta t sin(\theta_k) \\ 
    \theta_{k+1} = \theta_{k} + \omega_k\Delta t
\end{gather}  
where $(x_{k}, y_{k}, \theta_{k})$ represents the position and orientation of the robot at time $k$,  $v_k$ and $\omega_k$ are respectively the linear and angular velocities of the robot at time $k$, and $\Delta t$ is the sampling period. The possible linear and angular velocities are limited to 
\begin{gather}
    V_s = \{ (v, \omega) \mid v \in [v_{min}, v_{max}] \omega \in [\omega_{min}, \omega_{max}]\}.
    \label{eq:dwa1}
\end{gather}

Moreover, the velocities have to guarantee that the robot does not collide with the nearest obstacle. Let $dis(v, \omega)$ be the distance to the closest obstacle on the corresponding curvature created by a set of velocities $(v, \omega)$, and $(\dot{v_b}, \dot{\omega_b})$ are the accelerations for breakage. The set of admissible velocities is defined as
\begin{gather}
    V_a \{(v, \omega) | v \leq \sqrt{2\cdot dis(v, \omega)\cdot \dot{v_b}} \wedge  \omega \leq \sqrt{2\cdot dis(v, \omega)\cdot \dot{\omega_b}}
    \label{eq:dwa2}
\end{gather}

Considering the limited accelerations of the motors, the search space is reduced to a dynamic window containing available velocities in the next time interval. Let $(\dot{v}, \dot{\omega})$ are the accelerations being applied. The dynamic window $V_d$ is defined as
\begin{gather}
    V_d = \{ (v, \omega) \mid v \in [v-\dot{v}t, v+\dot{v}t] \wedge \omega \in [\omega-\dot{\omega}t, \omega+\dot{\omega}t]
    \label{eq:dwa3}
\end{gather}

From (\ref{eq:dwa1}) - (\ref{eq:dwa3}), the range of velocities can be restricted to
\begin{gather}
    V = V_s \cap V_a \cap V_d.
\end{gather}

In the next step, a set of $(v, \omega)$ is determined from $V$ by maximizing the objective function $G$ considering factors such as target heading, clearance to the obstacle, and velocity as the following:
\begin{gather}
    G = \alpha \cdot heading(v, \omega) + \beta \cdot dist(v, \omega) + \gamma \cdot velocity(v, \omega)  
\end{gather} where $\alpha, \beta, \gamma$ are positive constants.
The target heading function is defined as \begin{gather}
    heading(v, \omega) = 180 - \theta_g,
\end{gather} where $\theta_g$ is the angle between the robot's heading in the predicted position corresponding to a set of $(v, \omega)$ and the end goal's direction. This function measures the misalignment between the robot heading and the goal and thus is maximized when the robot moves directly to the goal. Function $dist(v, \omega)$ is the distance of the robot trajectory corresponding to velocities to the nearest obstacle. When encountering no obstacle, the function is set to a large constant and thus maximizes the objective function $G$. Function $velocity(v, \omega)$ is a projection of the vectors to the translation vector $v$. This function prioritizes the fastest velocity of the robot moving towards the goal. The optimal velocities $(v^*, \omega^*)$ found by maximizing the objective function $G(v,\omega)$ are finally applied to the robot to drive it in the next update.

\section{Results and discussion}
We have conducted a number of experiments to evaluate the human state estimation algorithm, the fusion algorithm, and the social-aware motion planner with details as follows.


\subsection{Human state estimation evaluation}
In the first experiment, we evaluate the performance of the human detection algorithm using LiDAR. Figure \ref{fighumanlidar} shows the detection result of the scenario in which a person standing in front of the robot. It can be seen that the joint leg tracker algorithm can correctly detect the person. However, some scan points of objects are also sometimes classified as human due to reflection noise leading to two people being detected, as shown in Fig. \ref{fig:lidar_error}. In contrast, the detection algorithm using the camera can reliably isolate the person from the background, as shown in Fig. \ref{fig:depth_camera}. This allows the fusion algorithm to correct the uncertainty in human detection using LiDAR. 
\begin{figure}[h]
\centering
\begin{subfigure}[t]{0.48\linewidth}
    \centering
    \includegraphics[width=1\textwidth]{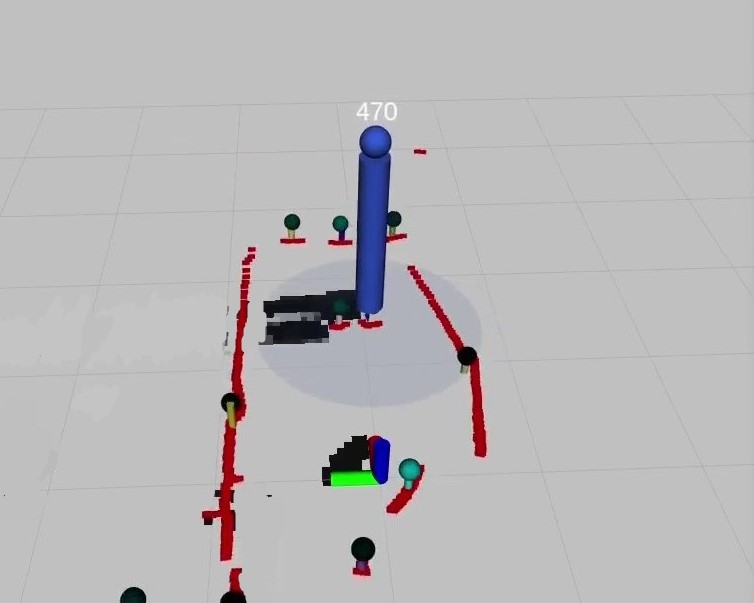}
    \caption{}
    \label{fig:lidar_normal}
\end{subfigure}
\hspace{0.05cm}
\begin{subfigure}[t]{0.48\linewidth} 
    \centering
    \includegraphics[width=1\textwidth]{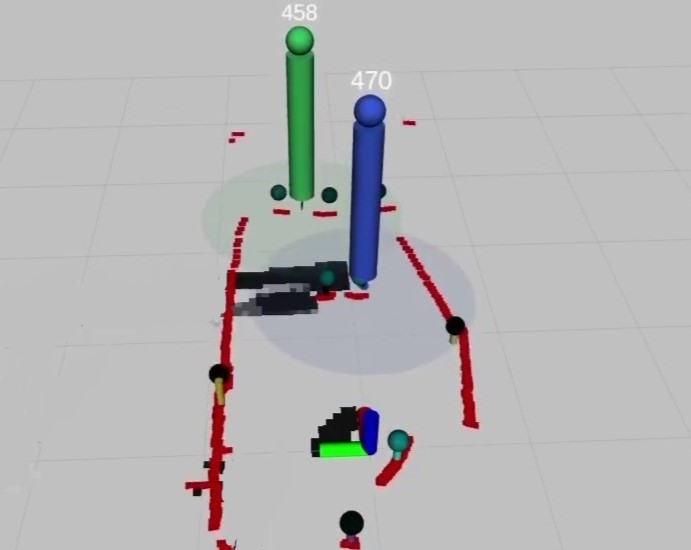}
    \caption{}
    \label{fig:lidar_error}
\end{subfigure}
\caption{Human state estimation by LiDAR: (a) correct detection of a person standing in front of the robot; (b)  wrong detection of two persons due to reflection noise}
\label{fighumanlidar}
\end{figure} 

To further evaluate the performance of the human state estimation algorithm on the RGB-D camera, we set up an experiment in which a person stands at different locations within the workspace of the robot. The mean and variance of the estimated positions are then computed as shown in Fig. \ref{fig:depth_human_error}, where the horizontal axis represents the real distance, and the vertical axes represent the mean (left) and variance (right) of the estimated positions. 
\begin{figure}[h]
    \centering
    \includegraphics[width=0.5\textwidth]{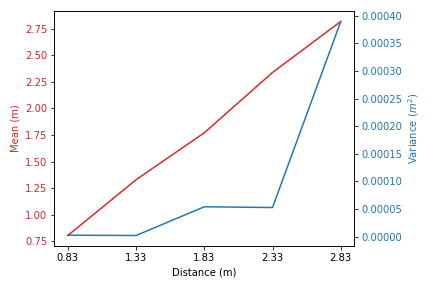}
    \caption{Mean and variance of human state estimated based on the RGB-D camera}
    \label{fig:depth_human_error}
\end{figure}
It can be seen that the mean is well aligned with the real distance implying an accurate estimation of the position. In addition, small values of the variance indicate the stability of the estimation. However, the accuracy and stability are decreased with respect to distance. This is due to the limitation of depth sensor technologies where the signal strength is dependant on the distance. This analysis suggests that the RGB-D camera is better used for detection from close to medium range.  
\begin{figure}[h]
    \centering
    \includegraphics[width=0.5\textwidth]{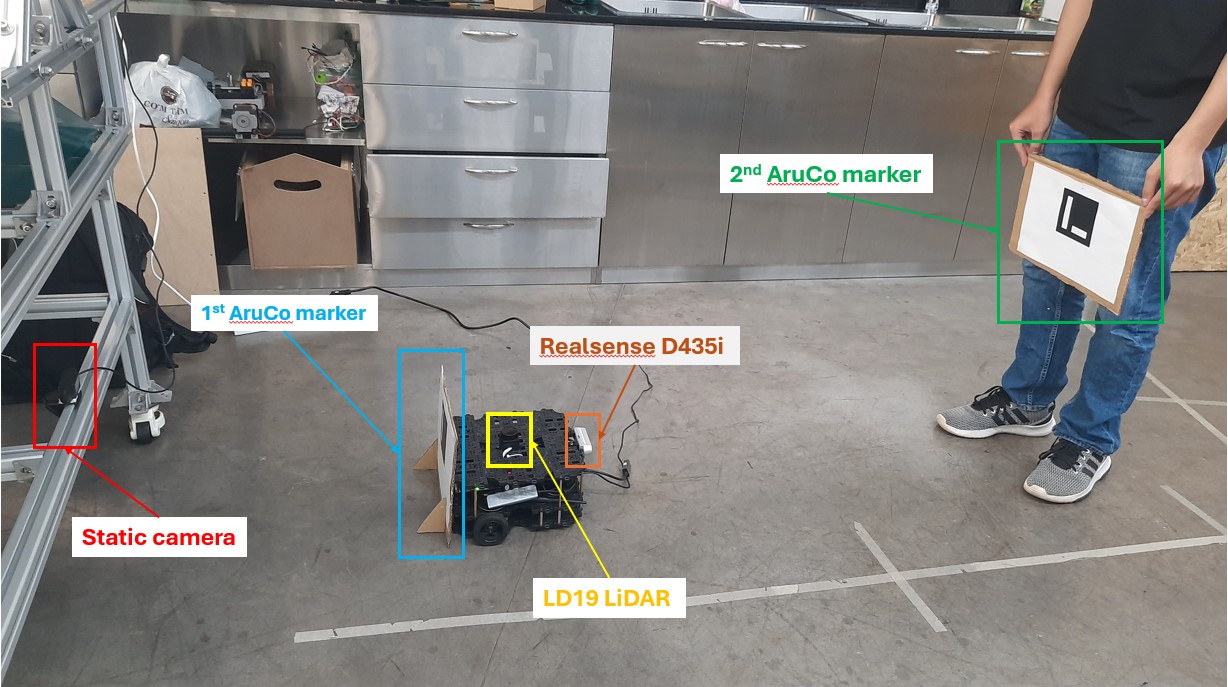}
    \caption{Experiment setup for the fusion algorithm where a static camera viewing two ArUco markers placed at the robot and the person}
    \label{fig:fusion}
\end{figure}

\subsection{Fusion algorithm evaluation}
To evaluate the fusion algorithm for LiDAR and the camera, we use ArUco markers attached to the people and the robot, and an external static monocular camera to track them, as shown in Fig. \ref{fig:fusion}. The position of the ArUco markers obtained by the external camera then will be compared with the position returned by the fusion algorithm. We evaluate the algorithms in two scenarios: (i) a person standing in front of the robot, and (ii) a person walking away from the robot along its $x$ direction. Figures \ref{fig:stand} and \ref{fig:walk} show the results. It can be seen that the estimated positions in both scenarios are more accurate and have less fluctuation compared to the estimation from the static camera. In addition, the mean square error (MSE) of the fused data is 5 times smaller than the MSE of the static camera data. The results thus confirm the accuracy and stability of the fusion algorithm.
\begin{figure}[h]
\centering
\begin{minipage}[t]{0.45\linewidth} 
    \centering
    \includegraphics[width=1\textwidth]{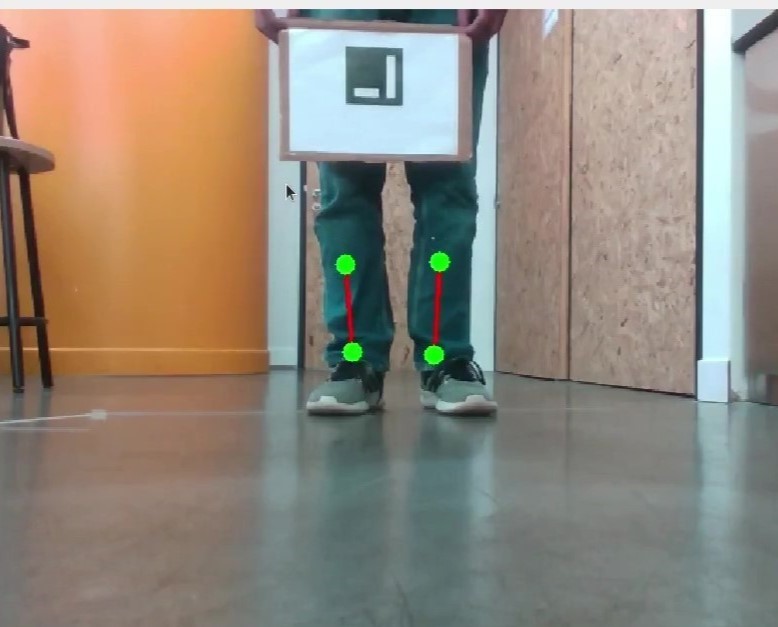}
    \caption{The FoV of the RGB-D camera showing the lower part of a person}
    \label{fig:depth_camera}
\end{minipage}
\hspace{0.05cm}
\begin{minipage}[t]{0.45\linewidth}
    \centering
    \includegraphics[width=1\textwidth]{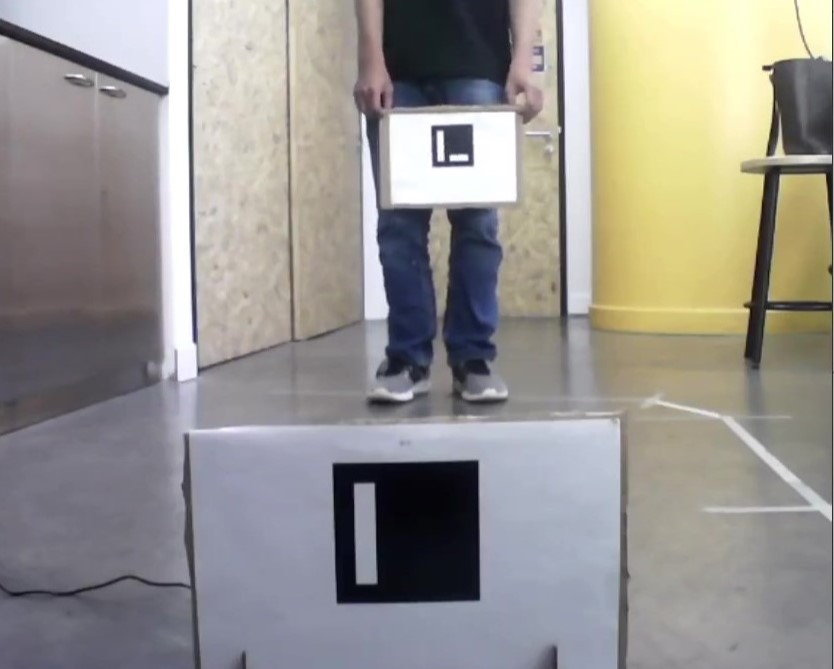}
    \caption{The FoV of the static camera showing the two ArUco markers}
    \label{fig:static_camera}
\end{minipage}

\end{figure} 
\begin{figure}[h]
    \centering
    \includegraphics[width=0.45\textwidth]{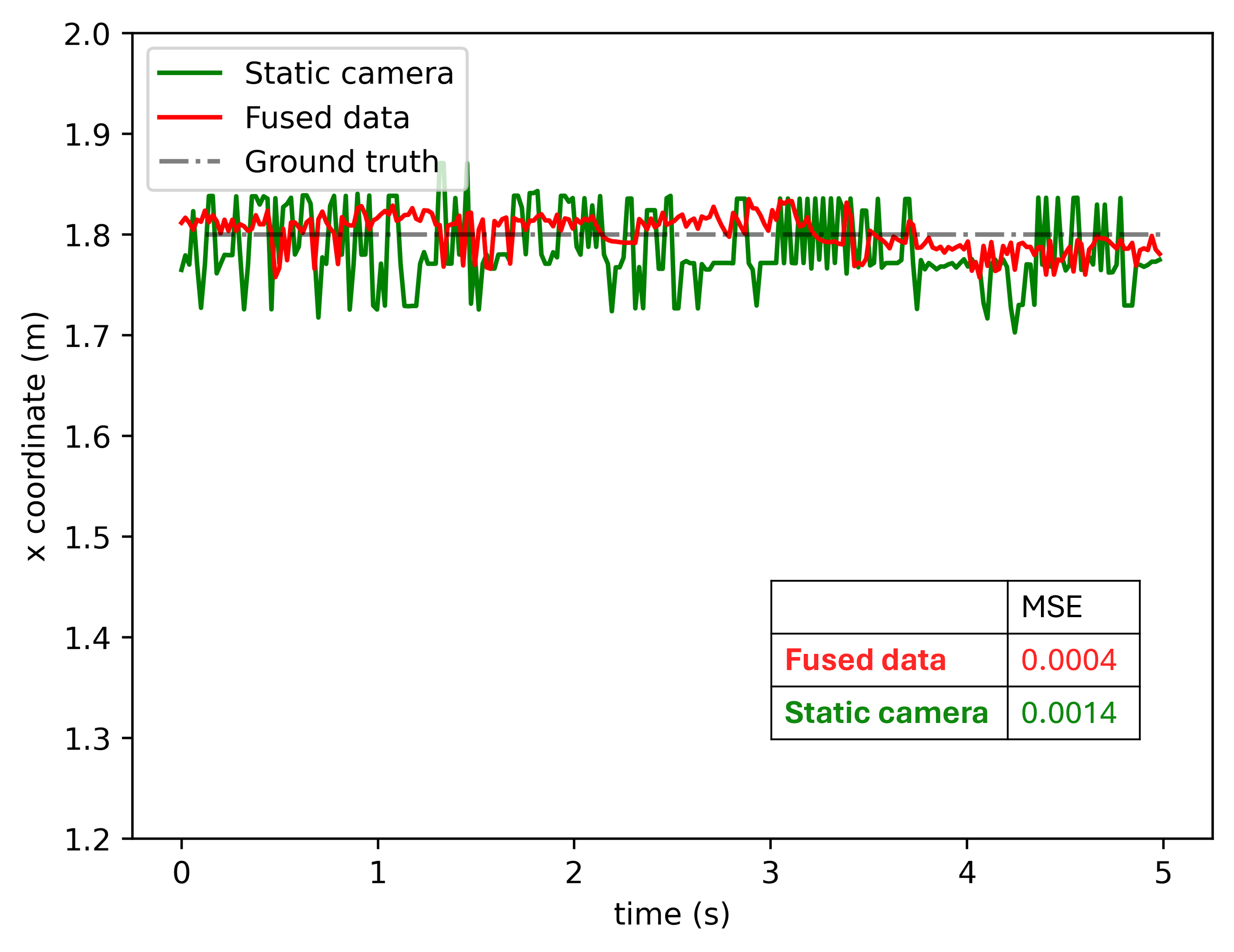}
    \caption{Estimated position and mean square error (MSE) when a person standing at 1.8 meters from the robot}
    \label{fig:stand}
\end{figure}
\begin{figure}[h]
    \centering
    \includegraphics[width=0.45\textwidth]{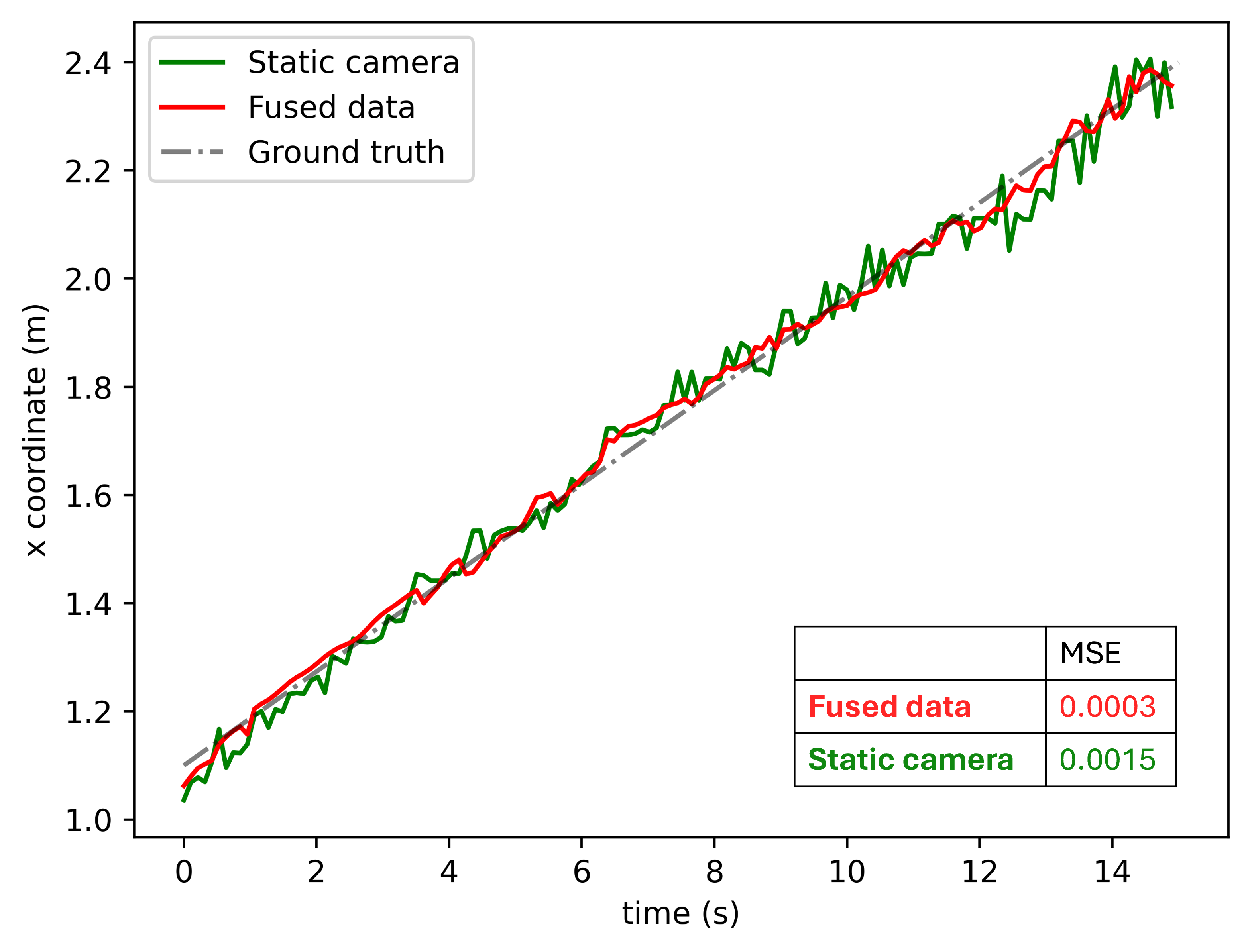}
    \caption{Estimated position and mean square error (MSE) when a person walking away 1 m to 2.4 m from the robot}
    \label{fig:walk}
\end{figure}
\subsection{Socially-aware motion planning evaluation}
We have set up experiments with two scenarios to evaluate the motion planning algorithm. The first scenario involves a robot moving in an environment having a standing person. The second scenario is more challenging in which a person attempts to interfere with the robot's trajectory. 

Figure \ref{fig:f1} shows the result for the first scenario. The social-aware motion planner successfully inscribes the region around the detected person with a red circle. The implemented DWA then reroutes the trajectory (the red line) to move further away from the person. In scenario 2, when a person interferes with the robot's trajectory, the inscribed region is described with an egg shape of which the cost is increased along the person's direction, as shown in Fig. \ref{fig:f2}. The DWA replanned the path in the opposite direction of the person's moving direction. This prevents potential collision and maintains comfort for the person. In this sense, the robot was aware of the human intention and adapted its operation to create a safe and cooperative environment.

\begin{figure}[H]
\centering
\begin{subfigure}[t]{0.48\linewidth}
    \centering
    \includegraphics[width=1\textwidth]{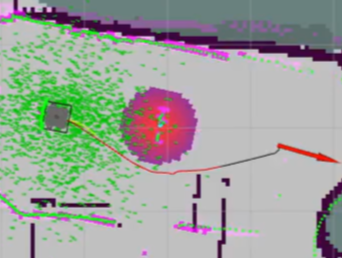}
    \caption{Scenario 1: a static person}
    \label{fig:f1}
\end{subfigure}
\hspace{0.05cm}
\begin{subfigure}[t]{0.48\linewidth} 
    \centering
    \includegraphics[width=1\textwidth]{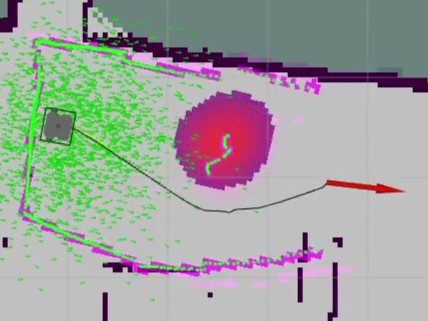}
    \caption{Scenario 2: a moving person}
    \label{fig:f2}
\end{subfigure}
\caption{Robot navigating through a person considering his personal space}
\label{fig:social}
\end{figure}

\section{Conclusion}
In this paper, we have introduced a motion planning method for socially aware navigation of service robots in dynamic environments. We proposed to use an asynchronous Kalman filter to fuse data from both LiDAR and the RGB-D camera for human state estimation. We then used an asymmetric Gaussian function to model the personal space of a person and incorporated it into a layer costmap to represent the environment. Finally, we implemented the dynamic window approach using the layered costmaps for social-aware navigation. Experimental results show that the fusion algorithm is accurate and robust in detecting and localizing humans' positions. The motion planning algorithm is able to guide the robot to avoid collisions with obstacles while maintaining proper distance from humans for their psychological comfort. 
\section*{Acknowledgement}
This work was partly supported by the IDP grant and TPBank STEM scholarship.

\bibliographystyle{ieeetr}
\bibliography{ref}

\begin{thebibliography}{10}

\bibitem{8011466}
X.-T. Truong and T.~D. Ngo, ``Toward socially aware robot navigation in dynamic and crowded environments: A proactive social motion model,'' {\em IEEE Transactions on Automation Science and Engineering}, vol.~14, no.~4, pp.~1743--1760, 2017.

\bibitem{rostami2019obstacle}
S.~M.~H. Rostami, A.~K. Sangaiah, J.~Wang, and X.~Liu, ``Obstacle avoidance of mobile robots using modified artificial potential field algorithm,'' {\em EURASIP Journal on Wireless Communications and Networking}, vol.~2019, no.~1, pp.~1--19, 2019.

\bibitem{fox1997dynamic}
D.~Fox, W.~Burgard, and S.~Thrun, ``The dynamic window approach to collision avoidance,'' {\em IEEE Robotics \& Automation Magazine}, vol.~4, no.~1, pp.~23--33, 1997.

\bibitem{lavalle2001randomized}
S.~M. LaValle and J.~J. Kuffner~Jr, ``Randomized kinodynamic planning,'' {\em The international journal of robotics research}, vol.~20, no.~5, pp.~378--400, 2001.

\bibitem{van2008reciprocal}
J.~Van~den Berg, M.~Lin, and D.~Manocha, ``Reciprocal velocity obstacles for real-time multi-agent navigation,'' in {\em 2008 IEEE international conference on robotics and automation}, pp.~1928--1935, Ieee, 2008.

\bibitem{chen2017socially}
Y.~F. Chen, M.~Everett, M.~Liu, and J.~P. How, ``Socially aware motion planning with deep reinforcement learning,'' in {\em 2017 IEEE/RSJ International Conference on Intelligent Robots and Systems (IROS)}, pp.~1343--1350, IEEE, 2017.

\bibitem{kirby2010social}
R.~Kirby, {\em Social robot navigation}.
\newblock Carnegie Mellon University, 2010.

\bibitem{papadakisadaptive}
P.~Papadakis, P.~Rives, and A.~Spalanzani, ``Adaptive spacing in human-robot interactions. in 2014 ieee,'' in {\em RSJ International Conference on Intelligent Robots and Systems}, pp.~2627--2632.

\bibitem{papadakis2013social}
P.~Papadakis, A.~Spalanzani, and C.~Laugier, ``Social mapping of human-populated environments by implicit function learning,'' in {\em 2013 IEEE/RSJ International Conference on Intelligent Robots and Systems}, pp.~1701--1706, IEEE, 2013.

\bibitem{truong2016dynamic}
X.-T. Truong and T.-D. Ngo, ``Dynamic social zone based mobile robot navigation for human comfortable safety in social environments,'' {\em International Journal of Social Robotics}, vol.~8, pp.~663--684, 2016.

\bibitem{2043184}
X.~Lu, H.~Woo, A.~Faragasso, A.~Yamashita, and H.~Asama, ``Socially aware robot navigation in crowds via deep reinforcement learning with resilient reward functions,'' {\em Advanced Robotics}, vol.~36, no.~8, pp.~388--403, 2022.

\bibitem{nakamori2018multiple}
Y.~Nakamori, Y.~Hiroi, and A.~Ito, ``Multiple player detection and tracking method using a laser range finder for a robot that plays with human,'' {\em ROBOMECH Journal}, vol.~5, pp.~1--15, 2018.

\bibitem{liu2016people}
H.~Liu, J.~Luo, P.~Wu, S.~Xie, and H.~Li, ``People detection and tracking using rgb-d cameras for mobile robots,'' {\em International Journal of Advanced Robotic Systems}, vol.~13, no.~5, p.~1729881416657746, 2016.

\bibitem{sarmento2024fusion}
J.~Sarmento, F.~Neves~dos Santos, A.~Silva~Aguiar, V.~Filipe, and A.~Valente, ``Fusion of time-of-flight based sensors with monocular cameras for a robotic person follower,'' {\em Journal of Intelligent \& Robotic Systems}, vol.~110, no.~1, pp.~1--14, 2024.

\bibitem{chebotareva2020person}
E.~Chebotareva, R.~Safin, K.-H. Hsia, A.~Carballo, and E.~Magid, ``Person-following algorithm based on laser range finder and monocular camera data fusion for a wheeled autonomous mobile robot,'' in {\em International Conference on Interactive Collaborative Robotics}, pp.~21--33, Springer, 2020.

\bibitem{lu2014layered}
D.~V. Lu, D.~Hershberger, and W.~D. Smart, ``Layered costmaps for context-sensitive navigation,'' in {\em 2014 IEEE/RSJ International Conference on Intelligent Robots and Systems}, pp.~709--715, IEEE, 2014.

\bibitem{leigh2015person}
A.~Leigh, J.~Pineau, N.~Olmedo, and H.~Zhang, ``Person tracking and following with 2d laser scanners,'' in {\em 2015 IEEE international conference on robotics and automation (ICRA)}, pp.~726--733, IEEE, 2015.

\bibitem{movenet}
Google, ``Movenet: Ultra fast and accurate pose detection model,'' Google Research, 2021.

\end{thebibliography}
\end{document}